\documentclass[letterpaper, 10 pt, conference]{ieeeconf}  

\IEEEoverridecommandlockouts                              

\usepackage{array}
\usepackage{graphicx}
\usepackage{amsmath}
\usepackage{amssymb}
\usepackage{booktabs}
\usepackage{hyperref}
\usepackage{multirow}
\usepackage{algorithmic, algorithm}
\usepackage{comment}
\usepackage{subfig}
\usepackage[compact]{titlesec}
\titlespacing{\subsection}{0pt}{*0}{*0}
\titlespacing{\subsubsection}{0pt}{*0}{*0}
\usepackage[font=footnotesize,skip=7.5pt,belowskip=-10pt]{caption}
\usepackage{tabularx,booktabs}
\newcolumntype{C}{>{\centering\arraybackslash}X} 
\title{\LARGE \bf
Real-Time Unified Trajectory Planning and Optimal Control for Urban Autonomous Driving Under Static and Dynamic Obstacle Constraints}

\author{Rowan Dempster$^*$, Mohammad~Al-Sharman, Derek Rayside, and William Melek
\thanks{R. Dempster, M. Al-Sharman, and D. Rayside are with the Department of Electrical and Computer Engineering, University of Waterloo. W. Melek is with the Department of Mechanical and Mechatronics Engineering, University of Waterloo. All authors are with the WATonomous lab, University of Waterloo, ON, N2L3G1, Canada. $^*$Corresponding author: { \tt\small r2dempst@watonomous.ca}}%
}

\begin{document}

\maketitle
\thispagestyle{empty}
\pagestyle{empty}

\begin{abstract}

Trajectory planning and control have historically been separated into two modules in automated driving stacks. Trajectory planning focuses on higher-level tasks like avoiding obstacles and staying on the road surface, whereas the controller tries its best to follow an ever changing reference trajectory. We argue that this separation is (1) flawed due to the mismatch between planned trajectories and what the controller can feasibly execute, and (2) unnecessary due to the flexibility of the model predictive control (MPC) paradigm. Instead, in this paper, we present a unified MPC-based trajectory planning and control scheme that guarantees feasibility with respect to road boundaries, the static and dynamic environment, and enforces passenger comfort constraints. The scheme is evaluated rigorously in a variety of scenarios focused on proving the effectiveness of the optimal control problem (OCP) design and real-time solution methods. The prototype code will be released at \href{https://github.com/WATonomous/control}{github.com/WATonomous/control}.
\end{abstract}

\section{INTRODUCTION}

In the SAE AutoDrive Challenge \cite{sae}, the Society of Automotive Engineers (SAE) and General Motors (GM) set forth guidelines for vehicle dynamics metrics that should be obeyed by autonomous vehicles (AVs) to ensure comfortable and safe urban driving. These guidelines include limits on longitudinal acceleration and jerk, as well as lateral acceleration in the vehicle's body frame. The guidelines are to be adhered to as the AV accomplishes the dynamic driving task (DDT). Generally, the DDT can be seen as progressing towards a goal state in a feasible and efficient manner. In the context of urban driving, feasible means without collisions and while obeying traffic rules, and efficient means operating near the speed limit. 

\subsection{\textbf{Motivation}}

There are several papers in the literature that address the real-time obstacle avoidance problem using optimal control techniques. However, all have drawbacks that make them ill-suited for the problem presented above. The authors of \cite{autogen} apply a nonlinear MPC method, similar to the one presented in Section \ref{sec:ocp_formulation}, to the obstacle avoidance problem. However, the controller must be instructed by the perception system on which direction to avoid the obstacle (left or right). This assumption is unsatisfactory because the directional decision in itself needs to consider the dynamics of the vehicle to know if a maneuver to the left or right of the obstacle is most optimal. The authors of \cite{pointmodel1} take an interesting approach towards integrated obstacle avoidance by representing obstacles as a potential field in the cost formulation. However, this paper, as well as \cite{pointmodel2}, \cite{wang2018predictive}, and \cite{jiang2016obstacle} have the issue that the avoidance maneuver controller (which they separate from the lower level actuator controller) is based on a particle representation of the vehicle. Therefore, these approaches have the same issue as using a higher level planner: the planned states generated by the lower fidelity planning model may not be dynamically feasible and therefore are unsafe.

\begin{figure}[t]
\centering
\includegraphics[width=0.48\textwidth]{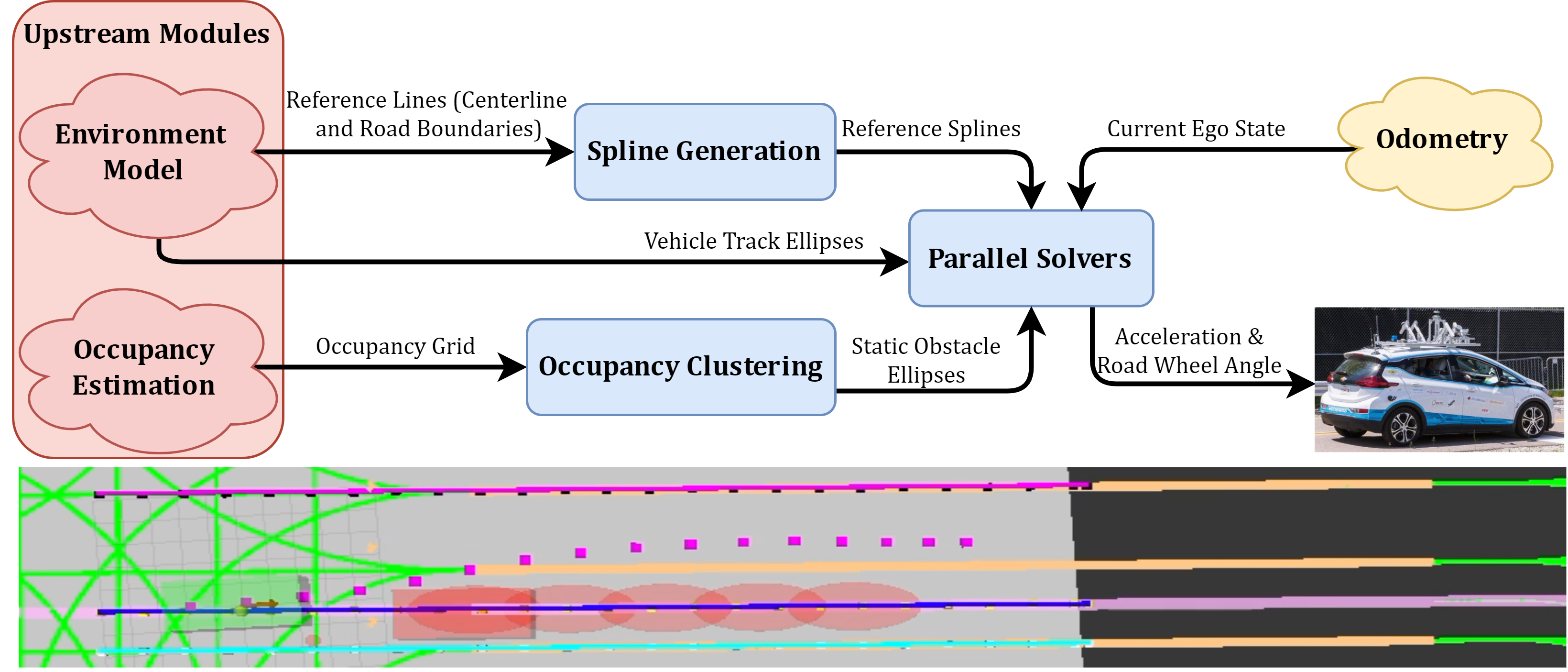}
\caption{The motion planning data pipeline discussed in Section \ref{sec:background}.} 

\label{fig:control:inputs}
\end{figure}

In \cite{cubicavoid}, a cubic polynomial is utilized to describe the lateral deviation from the reference line. Similarly, the authors of \cite{thesis} and \cite{li2019dynamic} use a sigmoid function. However, none of these papers provide an analysis of why these functions were chosen in the context of their OCP objective function. In contrast, the scheme proposed here avoids obstacles in a manner that is consistent with its OCP objective function because the obstacles themselves are part of the OCP formulation. 

In \cite{franco2019short}, an MPC-based technique for short-term path planning among multiple moving objects is presented. Experiments are carried out in simulation and present promising results in a variety of scenarios. However, the work depends on a linearized bicycle model assumption, which may not be a valid assumption during avoidance maneuvers that require large road wheel angles. The authors also assume that all obstacle information is known before system start-up and that the operating environment is a straight road, which makes their approach not applicable to a real-world setting.

\subsection{\textbf{Contributions}}

This paper addresses the issues presented above by introducing a novel OCP formulation of the DDT and an accompanying MPC solution. The main contributions in this regard are:
\begin{enumerate}
    \item The system model is designed to align closely with physical platforms and to allow for computation of comfort metrics and constraints, enabling the OCP to be formulated to abide by the SAE guidelines for passenger comfort.
    \item Constraints are added to the OCP formulation that guarantee feasibility of control actions with respect to road boundaries, static obstacles, and dynamic vehicles. In this way, trajectory planning can be executed simultaneously within the controller, and the final control actions are guaranteed to be dynamically feasible. 
    \item The controller operates in real-time by employing a novel parallel-solver method that allows for a warm-started ``online" solver, while employing a parallel ``exploration" solver which serves to break out of local warm-starting local minima.
\end{enumerate}

The remainder of this paper is organized as follows: Section \ref{sec:background} presents necessary background information on routing, reference line parameterization, and obstacle representations. Section \ref{sec:ocp_formulation} presents the OCP formulation, including the system model, objective function, road boundary constraints, and obstacle constraints. Section \ref{sec:mpc_solution} discusses the MPC solution to the OCP, including how the OCP is expressed as a nonlinear program (NLP), and the novel parallel-solver method for real-time performance. Experimental results are then presented and analyzed in Section \ref{sec:results}. Conclusions are discussed in Section \ref{sec:conclusion}.

\section{BACKGROUND}
\label{sec:background}

In order to properly understand the motion planning scheme in which the proposed controller fits, some background information is necessary. We first note the information given to the motion planning scheme:
\begin{enumerate}
    \item A high definition (HD) lane map.
    \item The desired goal position on the map.
    \item A discretized occupancy grid representation of the static environment.
    \item Non-ego dynamic vehicle tracks that describe the position and longitudinal velocity of the vehicle.
\end{enumerate}

\subsection{\textbf{Reference Spline Creation}}

In order to transform this information into reference signals that the controller can follow, we first need to find a lane-level route that describes how the vehicle can proceed from its current state to the goal state. This step is called global planning and is done using Dijkstra's search algorithm over the routing graph constructed from the HD map \cite{dempster2022drg}. 

However, a sequence of lanes is not yet an admissible reference signal for the controller. In order to generate a continuous and differentiable reference signal, a spline is fit to the centers of the lanes that make up the global route. In brief, CasADi's Interior Point Optimizer (IPOPT) \cite{wachter2006implementation} is used to find the spline coefficients that minimize squared error to the lane centers, see \cite{thesis} for details. 

An important note is that a spline  has a limited capacity to express the complex lane topology which may appear over a long route, incurring high squared error. To solve this problem, a sliding window is applied over the entire lane route, allowing the spline to accurately capture the simpler topology of the lanes in the local window. 

The same method is applied to generate splines for the left and right boundaries of the driveable surface. The lines that describes the driveable surface are determined using the routing rules stored in the HD map.


\subsection{\textbf{Occupied Space Representations}}

Aside from the centerline and road boundary splines, a reference for occupied space over time is also necessary to allow the controller to plan feasible trajectories. Two inputs are used to calculate this reference: An occupancy grid, and localized vehicle tracks from the tracker and environment model. The goal is to have a set of obstacles, and for each obstacle have a state trajectory that covers the MPC predictive horizon. The obstacle state representation used is a 2D ellipse, which has 5 degrees of freedom: X and Y of the centroid, RX (longitudinal radius), RY (lateral radius) and $\theta$ (yaw). To transform an occupancy grid into a set of obstacle trajectories, the grid is first filtered to only contain information of obstacles that are on the driveable surface. Then, a region growing algorithm is run with a neighborhood radius of four cells to cluster the obstacles in the grid (see Alg. \ref{alg:control:region-grow}). Lastly, the Minimum Volume Enclosing Ellipsoid (MVEE) approximate iterative algorithm \cite{mveematlab} \cite{todd2007khachiyan} is used to fit an ellipse to each cluster found by the region growing algorithm (see Alg. \ref{alg:control:grid-to-ellipses}). Since the occupancy grid contains information about the static portions of the environment only, the occupancy ellipses are the same for each step in the MPC horizon. 

\begin{algorithm}[!t]
    
    \caption{Region Grow}
    \begin{algorithmic}[1]
    \renewcommand{\algorithmicrequire}{\textbf{Input:}}
    \renewcommand{\algorithmicensure}{\textbf{Output:}}
    \REQUIRE{occupancy \textit{grid}, row \textit{r}, col \textit{c}, radius \textit{rad}, visited 2D array \textit{visited}}
    \ENSURE{\textit{[N, 2] array}, \textit{N} is the number of cells in cluster}
    \STATE{Initialize seedList $\gets$ [(r,c)]}
    \STATE{Initialize cluster $\gets$ [(r,c)]} 
    \STATE{Initialize neighbors $\gets$ [n for c in product(range(-rad, rad + 1), repeat=2]) if n[0] != 0 or c[1] != 0]}
    \WHILE{len(seedList) $>$ 0}
        \STATE{Initialize currCell $\gets$ seedList.pop(0)}
        \STATE{visited[currCell[0]][currCell[1]] = 1}
        \FOR{neigh $\in$ neighbors}
            \STATE{Initialize tR $\gets$ currCell[0] + neigh[0]}
            \STATE{Initialize tC $\gets$ currCell[1] + neigh[1]}
            \IF{grid[tR][tC] == 1 and visited[tR][tC] == 0}
                \STATE{seedList.append((tR, tC))}
                \STATE{cluster.append((tR, tC))}
            \ENDIF
        \ENDFOR
    \ENDWHILE
    \RETURN{cluster}
    \end{algorithmic} 
    \label{alg:control:region-grow}
\end{algorithm}

\begin{algorithm}[!t]
    
    \caption{Grid To Ellipses}
    \begin{algorithmic}[1]
    \renewcommand{\algorithmicrequire}{\textbf{Input:}}
    \renewcommand{\algorithmicensure}{\textbf{Output:}}
    \REQUIRE{occupancy \textit{grid}, neighbor radius \textit{rad}}
    \ENSURE{\textit{[N, 5] array}, \textit{N} is the number of ellipses in grid}
    \STATE{Initialize ells $\gets$ []}
    \STATE{Initialize visited $\gets$ zeros\_like(grid)} 
    \FOR{r in range(grid.height)}
        \FOR{c in range(grid.width)}
            \IF{visited[r][c] $==$ 0 and grid[r][c] $==$ 1}
                \STATE{cluster = RegionGrow(grid, r, c, rad, visited)}
                \STATE{C, rx, ry, theta = MVEE(cluster)}
                \STATE{ells.append(C.x, C.y, rx, ry, theta)}
            \ENDIF
        \ENDFOR
    \ENDFOR
    \RETURN{ells}
    \end{algorithmic} 
    \label{alg:control:grid-to-ellipses}
\end{algorithm}


To generate the ellipse trajectories for the dynamic vehicle tracks, the environment model module \cite{dempster2022drg} first localizes each track in the lanelet it geometrically occupies. Then, based on the estimated longitudinal velocity (assumed constant) from the tracker, and the geometry of the upcoming lanelets, the ellipses are generated following Alg. \ref{alg:control:dynamic-pred}. This simple prediction scheme can be expanded to a more complex learning based technique, but this simple method suffices for controller design.

\begin{algorithm}[!t]
    
    \caption{Predicted Dynamic Vehicle Trajectory}
    \begin{algorithmic}[1]
    \renewcommand{\algorithmicrequire}{\textbf{Input:}}
    \renewcommand{\algorithmicensure}{\textbf{Output:}}
    \REQUIRE{\textit{envModel}, \textit{track}, MPC step \textit{S} (in seconds), MPC horizon \textit{M}}
    \ENSURE{\textit{[M, 5] array}, vehicle ellipse trajectory prediction}
    \STATE{path = envModel.localizeVehicle(track)}
    \STATE{startDist = toArcCoordinates(path, track.pos)}
    \STATE{traj = []}
    \FOR{i in range(M)}
        \STATE{lonDist = startDist + i * S * track.lonVel}
        \STATE{p1 = interpPointAtDistance(path, lonDist)}
        \STATE{p2 = interpPointAtDistance(path, lonDist + 0.1)}
        \STATE{theta = atan2(p2.y - p1.y, p2.x - p1.x)}
        \STATE{traj.append(\{ p1.x, p1.y, track.rx, track.ry, theta \})}
    \ENDFOR
    \RETURN{traj}
    \end{algorithmic} 
    \label{alg:control:dynamic-pred}
\end{algorithm}

In summary, the information given to the controller is (as shown in Fig. \ref{fig:control:inputs}):
\begin{enumerate}
    \item A spline (denoted $S_{ref}$) that describes a local window of the global route.
    \item Two splines (denoted $S_{left}$ and $S_{right}$) which describe a local left and right drivable surface boundary.
    \item A set of predicted trajectory ellipses (denoted $O$, where $O_1$ is the set of obstacles at step 1, and $O_{1,1}$ is the prediction for obstacle 1 at the 1st step in the MPC horizon).
\end{enumerate}

\section{OCP FORMULATION}
\label{sec:ocp_formulation}
This section describes the plant system to be controlled, the predictive model used in the proposed MPC controller, the objective function, and the constraints. 

\subsection{\textbf{System and Predictive Models}}
\label{sec:control:sys_pred_model}

\textit{Input Variables}: The plant speed is controlled by applying a longitudinal acceleration command, and yaw is controlled via commanded road wheel angle. Additional, an input variable referred to as \textit{path progress} $u_{\xi}$ controls how far along the reference spline the next state will progress. The usage of this path progress input in the context of path following is explained below.

\textit{State Variables}: The state vector was chosen in order to maintain the vehicle dynamic metrics that are necessary to compute the constraints presented below. Specifically, the state vector maintains the vehicle's inertial pose tuple $[x_X, x_Y, x_\psi]^T$ as well as longitudinal and forward body frame velocities (denoted $x_{v_x}$, and $x_{v_{fwd}}$, respectively). Additionally, a state variable referred to as the \textit{path integral} $x_{\Xi}$ is defined as the integration of the path progress input mentioned above. The path integral variable keeps track of how far along the reference route the ego is, and its functionality in terms of the objective function is explained below. 

\textit{Output Variables}: The system output variables are the odometry solutions generated by the Inertial Navigation System (INS). The INS directly measures inertial pose, as well as longitudinal and lateral velocities and accelerations in the body frame. The current path integral is estimated using a discretized linear search over the reference spline's parameter range.

\textit{Predictive Model}: To describe the evolution of the system over time, a nonlinear model is necessary due to the nonlinear behavior of vehicular systems under large road wheel angles \cite{bimodel}. In an urban driving setting, maneuvers that require large road wheel angles are common. This is in contrast with a highway driving setting that may only require a few degrees of road wheel angle and hence operate inside of a linear dynamics range. To this end, a nonlinear kinematic bicycle model is employed, as illustrated in \cite{daoud2022simultaneous,Al-Sharman2022} and described by Eqs. \ref{eq:control:bicyle1}, \ref{eq:control:bicyle2}:


\begin{subequations}
\begin{equation}
  \dot{x}_X = x_{v_{fwd}}cos(x_{\psi} + \beta)
\end{equation}    
\begin{equation}
  \dot{x}_Y = x_{v_{fwd}}sin(x_{\psi} + \beta)
\end{equation}
\begin{equation}
  \dot{x}_{\psi} = \frac{x_{v_{fwd}}cos(\beta)tan(u_{\delta})}{l_f + l_r}
\end{equation}
\begin{equation}
  \dot{x}_{v_{x}} = u_a
\end{equation}
\begin{equation}
  \dot{x}_{\Xi} = u_{\xi}
\end{equation}
\label{eq:control:bicyle1}
\end{subequations}
where 
\begin{equation}
  \beta = \arctan(\frac{l_r tan(u_{\delta})}{l_f + l_r})
 \label{eq:control:bicyle2}
\end{equation} 
is the slip angle of the vehicle at its center of gravity (CoG).


In summary, a nonlinear kinematic bicycle model was chosen as it is a simple predictive model that captures the nonlinear system dynamics for low-speed urban navigation \cite{bimodel}, and thus is the best trade-off between NLP computation time and accurate prediction.

\subsection{\textbf{Objective Function}}
The terms included in the OCP objective function are:
\begin{enumerate}
    \item Reference position error. Calculated as squared error between $x_X$, $x_Y$ and the reference spline at arclength $x_{\Xi}$.
    \item Distance of path parameter to route completion. Calculated as $(1 - x_{\Xi})$.
\end{enumerate}

Giving the final objective function:

\begin{equation}
\begin{aligned}
    J = \sum_{k=1}^{N} \quad & \left[ \begin{bmatrix} x^k_X \\ x^k_Y \end{bmatrix} - \begin{bmatrix} x^k_{ref} \\ y^k_{ref} \end{bmatrix} \right]_2 + Q(1 - x^k_\Xi)
  \end{aligned}
\end{equation}

where $Q$ is a positive number. Thus, the higher the $Q$ value, the more incentivized the ego will be to complete the route, even at the cost of deviating from the reference route if needed. The outcome of the tuning process of $Q$ was two separate weighting settings, one for nominal path following and another setting for when the controller is avoiding an obstacle. In the path following setting, a $Q=1$ weight is used which ensures accurate following of the reference spline, whereas in the obstacle avoidance setting (when the state is within 5 meters of an obstacle), $Q=1000$ is used to encourage the vehicle to deviate from the reference if necessary to avoid the obstacle.

These two simple terms embody the non-safety constraint portions of the DDT: Drive close to the desired route, and make efficient progress towards route completion. The rest of the DDT (the safety constraints) is implemented as hard constraints on the NLP and are covered below. 

\subsection{\textbf{Constraints}}
\label{sec:control:constraints}
All vehicular systems are non-holonomic and therefore an OCP aiming to control such a system must take the non-holonomic constraints into consideration. These non-holonomic constraints include the system model presented above, as well as further physical constraints like maximum steering angle. There are also constraints enforced for the comfort of the passenger as per the SAE guidelines, which include limits on longitudinal and lateral acceleration as well as jerk. Most importantly, there are the constraints that enforce the feasibility of the predicted vehicle trajectory: The vehicle must stay in free space, defined as inside the road boundaries and not in collision with any obstacle. A complete list of the OCP constraints can be found in Table \ref{tab:control:constraints}.

\begin{table*}
 \caption{Optimal Control Problem Constraints}
\label{tab:control:constraints}
\begin{tabularx}{\textwidth}{p{1.5cm} p{2.5cm} p{2cm} p{5cm} p{4.6cm}}

\toprule
Variable(s)     &  Constraint & Type & Reasoning & Calculation \\ 
\midrule

$\boldsymbol{x}_{k+1}$   & $f(\boldsymbol{x}_{k}, \boldsymbol{u}_k)$      &  Non-Holonomic & Dynamics of system described via constraints as per multiple shooting \cite{BOCK19841603}          &  Runge-Kutta (RK4). See Section \ref{sec:control:sys_pred_model} for system used to model $f$.  \\

$u_{\delta}$ & $-\frac{\pi}{4} \leq u_{\delta} \leq \frac{\pi}{4}$ & Non-Holonomic & Physical range of wheel shaft & N/A, obtained from fact sheet \\
\addlinespace

$x_{v_x}$ & $0 \leq x_{v_x} \leq v_{MAX}$ & Legal & Vehicle cannot exceed speed limit & N/A, supplied by HD Map data \\

\addlinespace

$x_{a_y}$   &  $-3.5 \leq x_{a_y} \leq 3.5$  &  Comfort & SAE Bounds on lateral acceleration          &    From \cite{bimodellat}: $x_{a_y} = \frac{x_{v_x}^2u_\delta}{l_f + l_r}$ \\

$\dot{x}_{a_x}$ & $-10 \leq \dot{x}_{a_x} \leq 15$ & Comfort & SAE Bounds on longitudinal jerk & $\dot{x}_{a_x} = \frac{u^k_a - x^{k-1}_{a_x}}{T}$ Where $T$ is the sample time \\

$u_a$ & $-3.5 \leq u_{a} \leq 3.5$ & Comfort & SAE Bounds on longitudinal acceleration & N/A, input variable \\

\addlinespace


$[x_X, x_Y]'$   &    $[x_X,  x_Y]'$ $\in  S^{drive}$   & Feasibility &  The vehicle must be on the driveable surface        &  See Sec \ref{sec:control:road_bound}\\ 



$[x_X, x_Y]'$   &  $[x_X, x_Y]'$ $\not\in  \chi^{occ}$   & Feasibility &  The vehicle can only occupy free space          &  See Sec \ref{sec:control:obs_avoid}\\ 

$x_\Xi$ & $0 \leq x_\Xi \leq 1$ & Feasibility & Vehicle required to stop at end of route & N/A, state variable \\

\bottomrule
\end{tabularx}
\end{table*}


\subsubsection{Road Boundary Enforcement}
\label{sec:control:road_bound}

There are multiple ways to enforce that a 2D point $[x_X, x_Y]'$, i.e. vehicle position, must be inside a set of two splines. The method we found worked best was a ``sidedness" test, enforcing that $[x_X, x_Y]'$ is to the right of the left bound, and to the left of the right bound. The first step in the formulation is to obtain a 2D bound vector $[b_0, b_1]'$ that we enforce the sidedness constraint with respect to. $b_0$ is calculated as the boundary spline at arclength $x_\Xi$ and $b_1$ is calculated as the boundary spline at arclength $x_\Xi$ + $la$ where $la$ is a small look-ahead (e.g. 0.01). Then, the sidedness constraint can be enforced as in Eq. \ref{eq:control:road_bound}

\begin{equation}
\label{eq:control:road_bound}
\begin{aligned}
    (x_X - b_0.x) * (b_1.y - b_0.y) - \\
    (x_Y - b_0.y) * (b_1.x - b_0.x) > 0
  \end{aligned}
\end{equation}

\subsubsection{Obstacle Avoidance Enforcement}
\label{sec:control:obs_avoid}

For each obstacle, and for each step in the MPC horizon, the vehicle's footprint cannot intersect with any obstacle ellipse. Eq. \ref{eq:control:in_ellipsoid} \textit{InEllipse(p, el)} implements the basic operation needed for such a constraint, testing if a 2D point $p$ is inside an ellipsoid $el$ (assuming that the point and the ellipsoid are in the same reference frame, a detail that is worked out in Alg. \ref{alg:control:obs_avoid}).

\begin{equation}
\label{eq:control:in_ellipsoid}
\begin{aligned}
    \frac{(p.x - el.x)}{el.rx}^2 + \frac{(p.y - el.y)}{el.ry}^2 < 1
  \end{aligned}
\end{equation}

Now all that is left is to call \textit{InEllipse(p, el)} for each ellipse we want to avoid at each step in the horizon as implemented in Alg. \ref{alg:control:obs_avoid}. Where the $expandFootprint$ subroutine simply returns 6 points around the border of the vehicle's footprint, and $ENFORCE$ enforces the enclosed constraint in the IPOPT solver. Note that here we are creating $6 * N * M$ constraints. This can be on the order of 100s of constraints for a nominal $N=15$ and $M < 10$.

\begin{algorithm}[!t]
    
    \caption{Obstacle Avoidance}
    \begin{algorithmic}[1]
    \renewcommand{\algorithmicrequire}{\textbf{Input:}}
    \renewcommand{\algorithmicensure}{\textbf{Output:}}
    \REQUIRE{Symbolic set of ellipse trajectories $O$, symbolic state trajectory $S$}
    \ENSURE{For each matching $O_i$, $S_i$ in the horizon, enforce that $S_i$ does not conflict with any object in $O_i$}
    \FOR{$O_i \in O$, $S_i \in S$}
        \FOR{$O_{i,j} \in O_i$}
            \STATE{th = $O_{i,j}.theta$}
            \STATE{rot = [cos(th), sin(th); -sin(th), cos(th)]}
            \STATE{el = \{rot * $O_{i,j}.cent$, $O_{i,j}.rx$, $O_{i,j}.ry$\}}
            \FOR{$S_{i,j} \in expandFootprint(S_i)$}
                \STATE{ENFORCE(! InEllipse(rot * $S_{i,j}.pos$, el))}
            \ENDFOR
        \ENDFOR
    \ENDFOR
    \end{algorithmic} 
    \label{alg:control:obs_avoid}
\end{algorithm}

\section{MPC SOLUTION}
\label{sec:mpc_solution}
To solve the OCP, nonlinear MPC (NMPC) was used. This decision was due to: (1) The high number of constraints used to enforce non-holonomic system dynamics, comfort, and feasibility, and (2) the nonlinear dynamics of the system.


In order to transform the OCP into a NLP, temporal discretization was applied to the system dynamics over a finite prediction horizon ($N=15$). Specifically, the Runge–Kutta method (RK4) was used with a time step of $T = 0.25s$. As proposed by Bock and Plitt in \cite{BOCK19841603}, multiple shooting was used to enforce the dynamics of the system using constraints, reducing the nonlinearity of the objective function especially in the latter steps of the prediction horizon. 

To solve the NLP, the interior point optimizer (IPOPT) \cite{wachter2006implementation} from the CasADi \cite{andersson2019casadi} software package was used. Computational concerns regarding real-time IPOPT solutions are discussed below. With the NLP solution obtained, receding horizon control is applied.



    

\subsection{\textbf{Real Time Performance}}
\label{sec:control:parallel_solvers}
Real-time performance of the controller is defined as the NLP solver operating faster than the sampling time of the controller (0.25s), i.e. greater than 5Hz. In order to achieve real-time performance, warm-starting the iterative solver with a ``decent" solution is essential, allowing the solver to reach an acceptable solution after only a small number of iterations. The warm-start used at time $t$ is normally the solution obtained by the solver at time $t-1$, following from the assumption that the parameters of the optimization change little from one solve to the next. However, over time this assumption may no longer be true, and the solver may become stuck in a series of warm-started local minima (see Section \ref{sec:control:res_parallel_solver} for an example). 

A method of ``breaking out" of this local minima is required, a way to search for a more global minima given more IPOPT iterations and a less biased warm-start. Our novel method is to run a parallel solver that consumes the same NLP as the original solver (which we will refer to from now on as the ``online" solver), but does not use warm-starting (initial decision variable assignments are all zero) and is allowed 10x the number of IPOPT iterations. We will refer to this new solver as the ``exploration" solver. 

These two solvers run asynchronously of each other, the online solver is used to control the vehicle as normal, and the exploration controller is used to simply ``suggest" new warm-starts that the online solver could use. Whenever the online solver is about to perform a new solve, it polls the exploration solver for its most recent solve. If the polled solution has a lower cost than the cost of the previous online solution, the exploration solution is used for the online solver's warm-start instead of the previous online solution. 


The result, viewed from a unified perspective, is a controller than can operate in real-time using a warm-starting scheme but also does not get stuck in local minima (see Section \ref{sec:control:res_parallel_solver} for results).


\section{EXPERIMENTS}
\label{sec:results}

\subsection{\textbf{Reference Line Following}}
\label{sec:control:res_ref_follow}

In this experiment, the only reference we have to follow is the lane center as the ego vehicle completes a right turn at a 4-way intersection. The desired behavior is to have low lateral error from the reference spline, keep the NLP solution time within the 0.25s sampling time, and obey the legal and comfort constraints. 



\textit{Qualitative results} of the controller completing the scenario are shown at:\\\url{https://youtu.be/2Gk3XQKlk38}.

\begin{figure}[h!]
\centering
\includegraphics[width=0.45\textwidth, height=120pt]{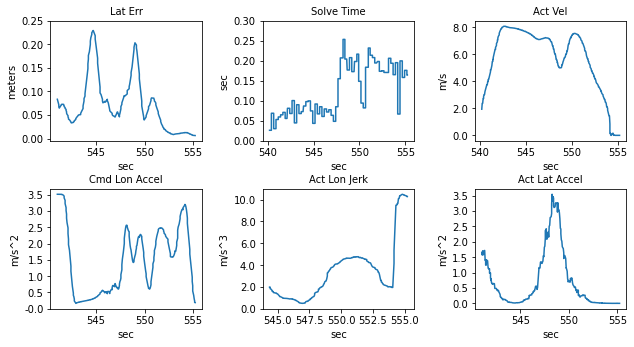}
\caption{Quantitative reference line tracking performance.}
\label{fig:control:res-1-1-normal}
\end{figure}

\textit{Quantitative results} are shown in Fig. \ref{fig:control:res-1-1-normal}. As seen in the figure, the lateral deviation stays below 25cm from the reference, well within the lane boundaries. The solution time of the NLP also stays below 0.25s, and the velocity stays close to 8m/s, except when necessary to slow down while taking the right turn to avoid passenger discomfort. The comfort metrics are all within the desired range outlined by SAE. 

\subsection{\textbf{Static Obstacle Avoidance}}
\label{sec:control:results_static_obs_avoid}

In this experiment, a new reference is introduced: static obstacles. The scenario and desired behavior is the same as in Section \ref{sec:control:res_ref_follow}, with the addition of a static obstacle course. 




\textit{Qualitative results} of the controller completing the scenario are shown at:\\\url{https://youtu.be/T83wHpZDfd0}.

\begin{figure}[h!]
\centering
\includegraphics[width=0.45\textwidth, height=120pt]{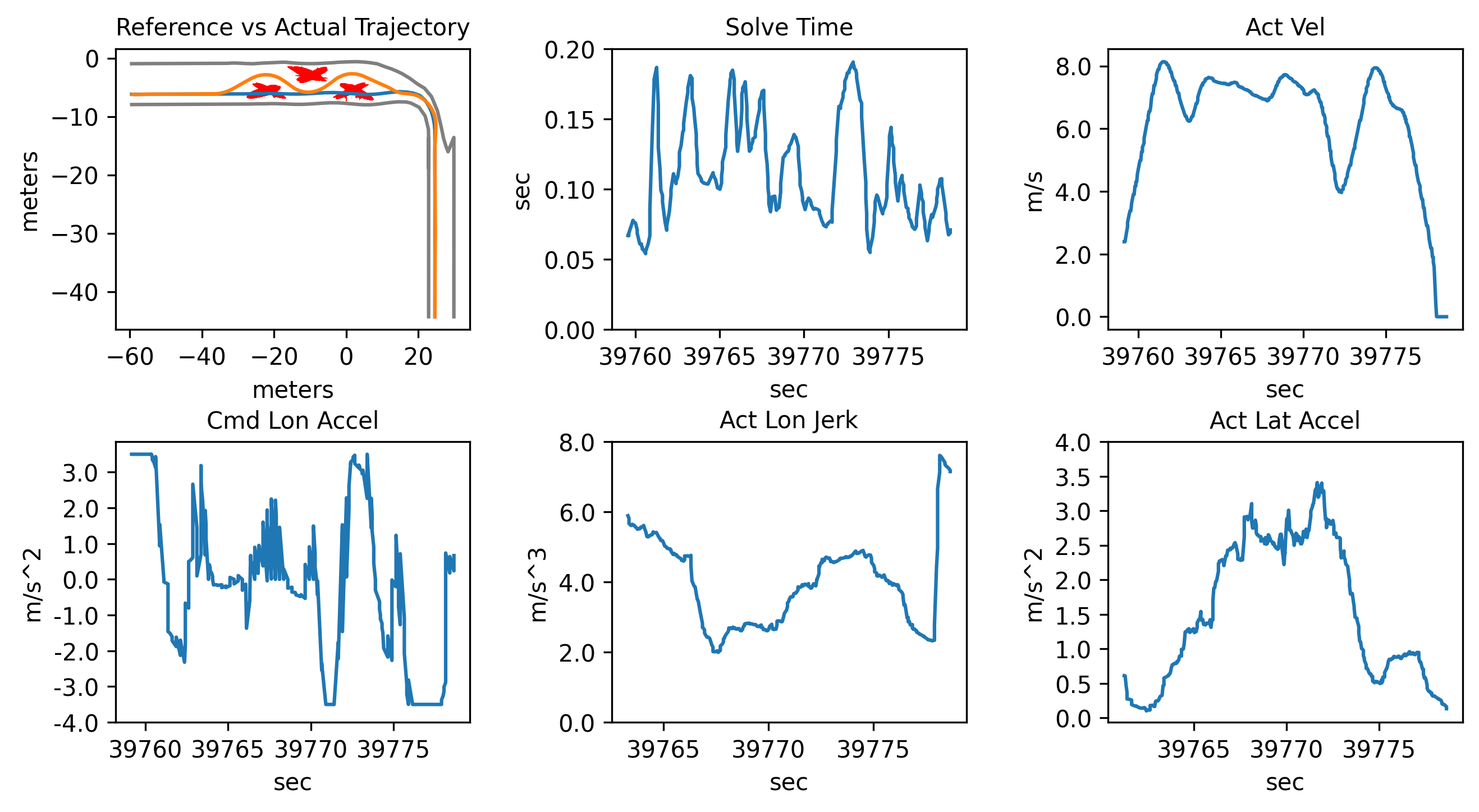}
\caption{Quantitative static obstacle avoidance performance.}
\label{fig:control:res_static_normal}
\end{figure}

\textit{Quantitative results} are shown in Fig. \ref{fig:control:res_static_normal}. As shown in the top left subplot, the reference (in blue) is tracked well when the controller is not performing an obstacle avoidance maneuver. The controller also successfully avoids the static obstacles (in red). Furthermore, the trajectory stays on the driveable surface (bounds in grey).

\subsection{\textbf{Road Boundary}}
To show the effectiveness of the road boundary constraints discussed in Section \ref{sec:control:road_bound}, they were removed in this experiment and the resulting vehicle behavior is discussed below. The same static obstacle course testing scenario was used to examine the results.




\textit{Qualitative results} of the no-road-boundary controller failing to complete the scenario are shown at:\\\url{https://youtu.be/YqpCCIzRw28}.

\begin{figure}[h!]
\centering
\includegraphics[width=0.45\textwidth, height=120pt]{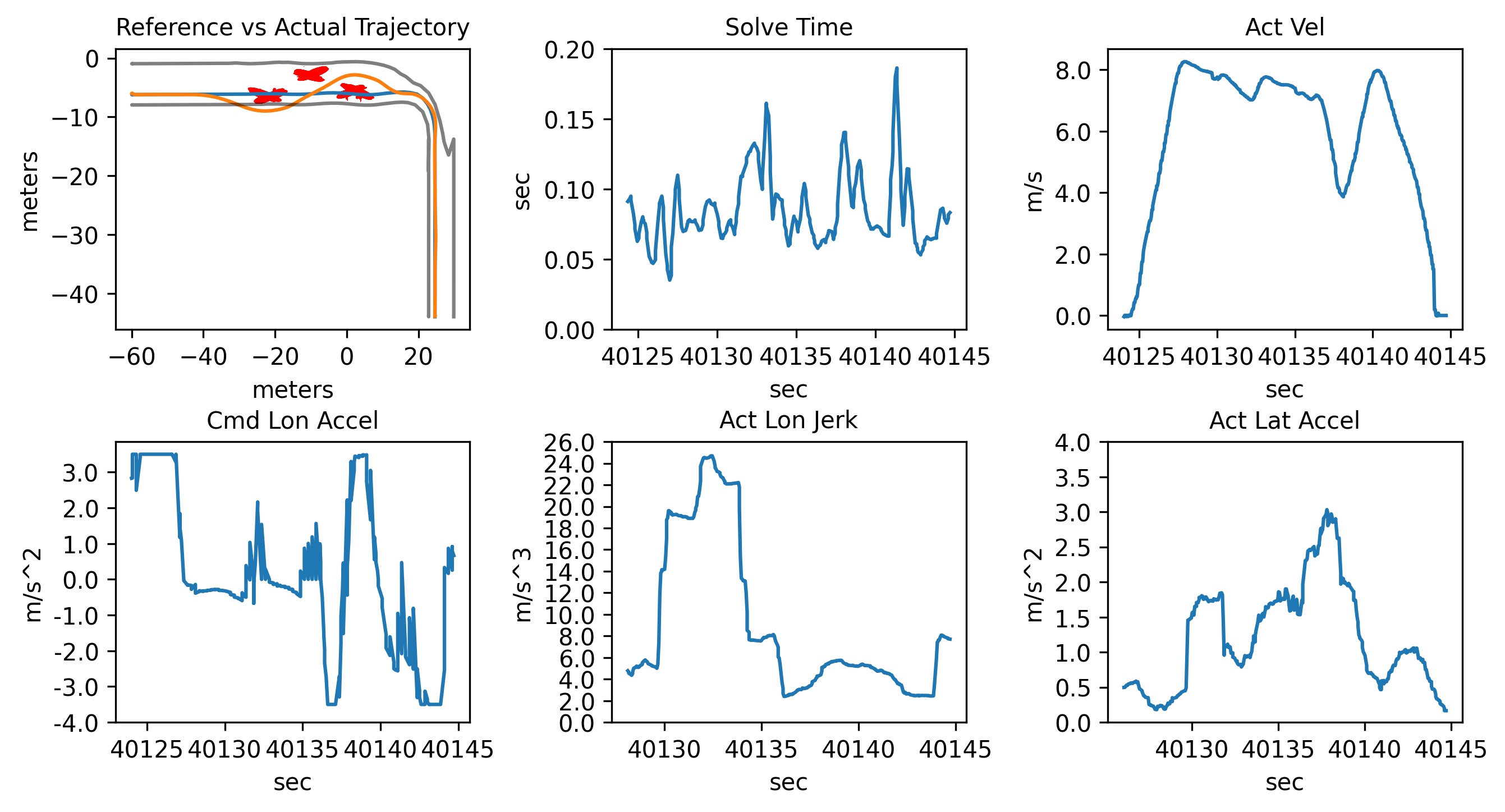}
\caption{Quantitative static obstacle avoidance performance of the no-road-boundary setting.}
\label{fig:control:res_no_road_bounds}
\end{figure}

\textit{Quantitative results} are shown in Fig. \ref{fig:control:res_no_road_bounds}. Without the road boundary constraints the controller may arbitrarily decide to avoid the obstacle on either side, possibly violating the road boundary traffic rule (see top left plot). This plot, when compared against the plot in Fig. \ref{fig:control:res_static_normal}, shows that the road boundary constraints specified in Section \ref{sec:control:road_bound} are necessary for correct driving behavior.

\subsection{\textbf{Parallel Solver}}
\label{sec:control:res_parallel_solver}
To show the effectiveness of the parallel solver technique discussed in Section \ref{sec:control:parallel_solvers}, it was removed (only the online solver is used) in this experiment. The same static obstacle course testing scenario was used to examine the results. The resulting vehicle behavior is discussed below.


\textit{Qualitative results} of the purely online solver failing to complete the scenario are shown at:\\\url{https://youtu.be/cLFCUO3-lMY}. Without the parallel solver the controller gets stuck in a local warm-starting minima and fails to avoid the first obstacle.



\begin{figure}[h!]
\centering
\includegraphics[width=0.45\textwidth, height=120pt]{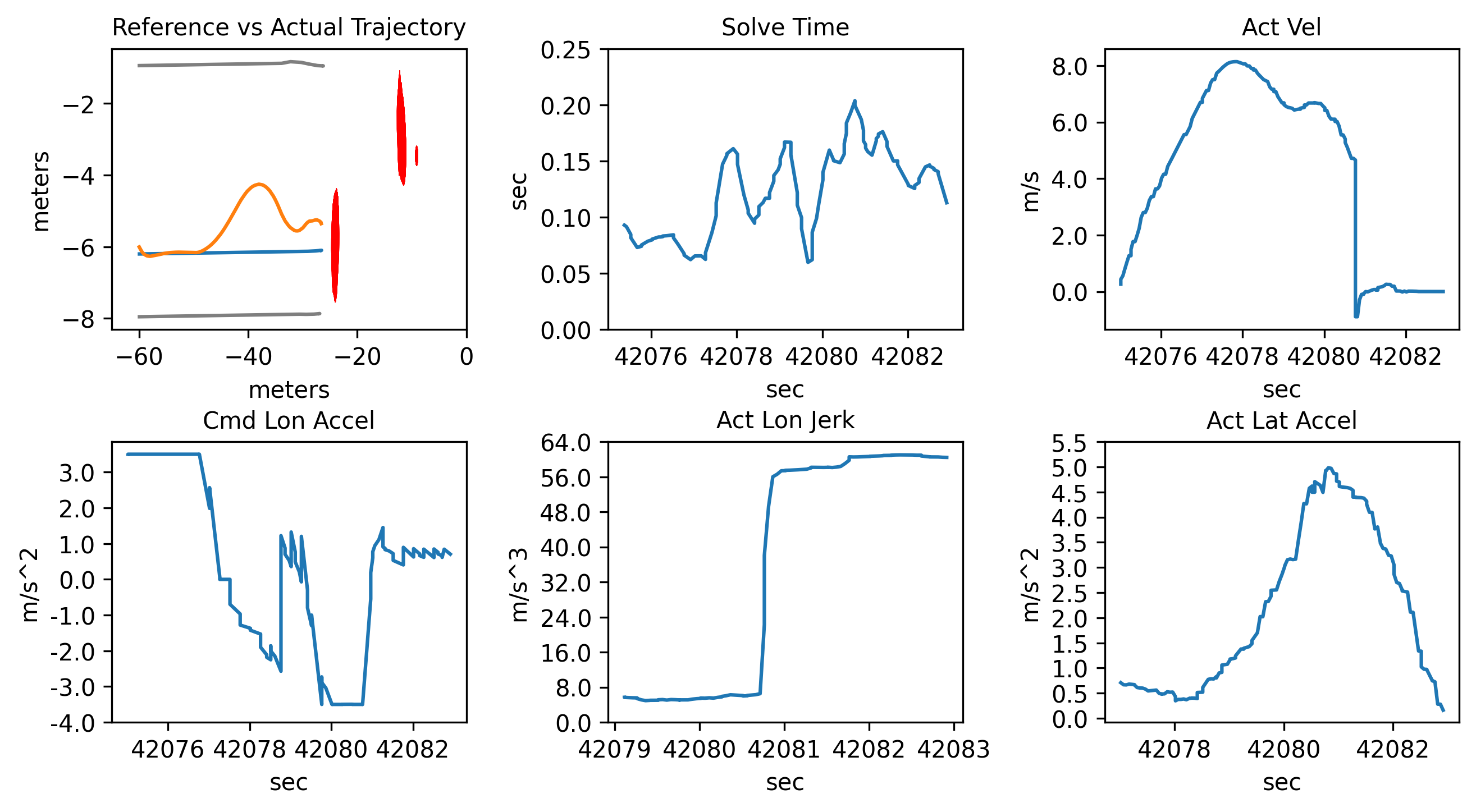}
\caption{Quantitative static obstacle avoidance performance of the no-parallel solver setting.}
\label{fig:control:res_no_parallel}
\end{figure}

\textit{Quantitative results} are shown in Fig. \ref{fig:control:res_no_parallel}. Without the parallel solver, the controller cannot find a trajectory around the first static obstacle on the course. The failure observed here compared to the success in Section \ref{sec:control:results_static_obs_avoid} is due to the missing exploration solver discussed in Section \ref{sec:control:parallel_solvers}.

\subsection{\textbf{Dynamic Obstacle Avoidance}}

In this experiment, we introduce a new reference: dynamic obstacles. The scenario used is a straight road with the ego obeying a speed limit of 15m/s. In front of the ego, there is a target vehicle traveling at 7.5m/s. In order to make efficient progress along the route, the ego vehicle should overtake the target vehicle in a comfortable and legal manner.


\textit{Qualitative results} of the controller completing the scenario are shown at:\\\url{https://youtu.be/oqjdudBFZ9E}.



\begin{figure}[h!]
\centering
\includegraphics[width=0.45\textwidth, height=120pt]{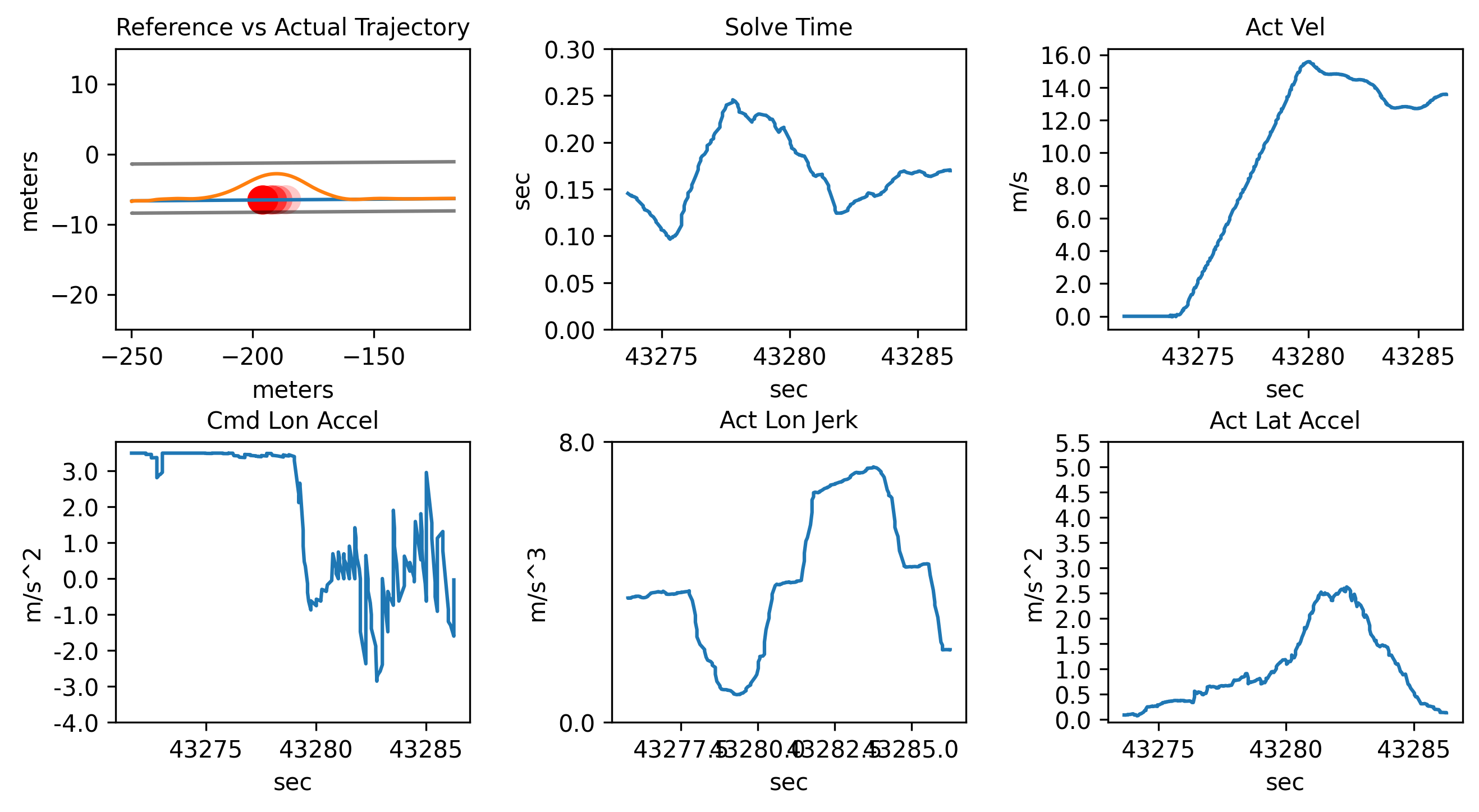}
\caption{Quantitative dynamic vehicle avoidance performance.}
\label{fig:control:res_dyn_avoid}
\end{figure}

\textit{Quantitative results} are shown in Fig. \ref{fig:control:res_dyn_avoid}. The reference is tracked well when the controller is not overtaking the target vehicle (see top left plot). The controller also successfully avoids the target vehicle (in semi-transparent red). Furthermore, the controller performs the maneuver efficiently (near 15m/s) and comfortably (SAE guidelines obeyed) while staying on the driveable surface.

\section{CONCLUSION}
\label{sec:conclusion}
We present a novel OCP formulation and MPC solution to the DDT that allows for unified trajectory planning and control in a feasibility guaranteed manner. The presented scheme is shown to have key advantages over previous DDT motion planning and execution schemes, including direct adherence to SAE passenger comfort guidelines, as well as free space and road boundary conscious control via the OCP constraints. Promising experimental results were achieved for obstacle avoidance, overtaking, and turning maneuvers. 

As for future research directions, incorporating learning techniques for more accurate predictions of non-ego vehicle trajectories will enhance the motion planning scheme\footnote{Note that the OCP design will not have to change, only the predictive accuracy of the inputs does.}. Furthermore, we plan to integrate the proposed motion planning and control framework with learning-based behavioral planners, e.g. \cite{Al-Sharman2022}, for developing feasible decision-making schemes in complex urban environments.

\section*{Acknowledgment}

This work was supported by NSERC CRD 537104-18, in partnership with General Motors Canada and the SAE AutoDrive Challenge.

{\small
\bibliographystyle{IEEEtran}
\bibliography{IEEEabrv, references}
}

\end{document}